\documentclass[conference]{IEEEtran}
\IEEEoverridecommandlockouts
\usepackage{cite}
\usepackage{amsmath,amssymb,amsfonts}
\usepackage{algorithmic}
\usepackage{graphicx}
\usepackage{textcomp}
\usepackage{xcolor}
\usepackage{tikz}

\def\BibTeX{{\rm B\kern-.05em{\sc i\kern-.025em b}\kern-.08em
    T\kern-.1667em\lower.7ex\hbox{E}\kern-.125emX}}
\begin{document}

\title{Evaluating the Impact of Adversarial Attacks on Traffic Sign Classification using the LISA Dataset}

\author{
\IEEEauthorblockN{Nabeyou Tadessa\textsuperscript{1} (Student)}
\IEEEauthorblockA{\textsuperscript{1}\textit{Department of Computer Engineering} \\
\textit{Benedict College} \\
Columbia, SC, USA \\
nabeyou.tadessa79@my.benedict.edu}
\and
\IEEEauthorblockN{Balaji Iyangar\textsuperscript{2} (Associate Professor)}
\IEEEauthorblockA{\textsuperscript{2}\textit{Department of Computer Science} \\
\textit{Benedict College} \\
Columbia, SC, USA \\
Balaji.Iyangar@benedict.edu}
\and
\IEEEauthorblockN{Mashrur Chowdhury\textsuperscript{3} (Professor)}
\IEEEauthorblockA{\textsuperscript{3}\textit{Department of Civil Engineering} \\
\textit{Clemson University} \\
Clemson, SC, USA \\
mac@clemson.edu}
}

\maketitle

\begin{abstract}
Adversarial attacks pose significant threats to machine learning models by introducing carefully crafted perturbations that cause misclassification. While prior work has primarily focused on MNIST and similar datasets, this paper investigates the vulnerability of traffic sign classifiers using the LISA Traffic Sign dataset. We train a convolutional neural network to classify 47 different traffic signs and evaluate its robustness against Fast Gradient Sign Method (FGSM) and Projected Gradient Descent (PGD) attacks. Our results show a sharp decline in classification accuracy as the perturbation magnitude increases, highlighting the model's susceptibility to adversarial examples. This study lays the groundwork for future exploration into defense mechanisms tailored for real-world traffic sign recognition systems.
\end{abstract}

\begin{IEEEkeywords}
Adversarial Attacks, Deep Learning, LISA Dataset, Traffic Sign Classification, FGSM, PGD, Robustness Evaluation
\end{IEEEkeywords}

\section{Introduction}
Deep learning models have achieved remarkable success in a wide range of applications, including image classification, natural language processing, and autonomous driving. Despite their impressive performance, these models are vulnerable to adversarial attacks, which are small, intentionally crafted perturbations that can cause the model to make incorrect predictions with high confidence. 

Research on adversarial robustness has focused primarily on standard benchmark datasets such as MNIST, CIFAR-10, and ImageNet. However, the real-world implications of adversarial attacks extend beyond simple datasets to safety-critical domains such as traffic sign recognition, where misclassifications can have severe consequences.

In this paper, we investigate adversarial vulnerabilities in a traffic sign classification system using the LISA Traffic Sign dataset. We train a convolutional neural network to recognize 47 different traffic sign categories and evaluate the model's robustness against two popular white-box attack methods: Fast Gradient Sign Method (FGSM) and Projected Gradient Descent (PGD). By analyzing how classification accuracy degrades under adversarial perturbations of varying magnitudes, we aim to better understand the risks of deploying deep learning models in real-world autonomous systems.

This work replicates the methodology of previous studies conducted on simpler datasets, but adapts the evaluation to a more complex and practically relevant dataset. The findings serve as a baseline for future research on defense mechanisms tailored to traffic sign recognition tasks.

\section{Related Work}
Adversarial attacks on machine learning models were first brought to attention by Szegedy et al. (2014), who showed that deep neural networks could be easily fooled by adding small, imperceptible perturbations to the input data, leading to misclassifications. Building on this, Goodfellow et al. (2015) introduced the Fast Gradient Sign Method (FGSM), a computationally efficient white-box attack that uses the model’s gradients to craft adversarial examples. Subsequently, Madry et al. (2018) proposed Projected Gradient Descent (PGD), a more powerful iterative variant of FGSM that identifies stronger adversarial inputs within a defined perturbation bound. These foundational works have spurred extensive research into adversarial robustness, including adversarial training, detection mechanisms, and certified defenses (Athalye et al., 2018; Cohen et al., 2019). However, most evaluations have centered on relatively simple and controlled datasets such as MNIST and CIFAR-10, limiting the generalizability of findings.

In contrast, complex, real-world datasets like the LISA Traffic Sign dataset—which includes traffic signs under varying illumination and environmental conditions—have received limited attention in adversarial robustness research. To address this gap, our work applies established attack methodologies to the LISA dataset, aiming to assess the vulnerability of traffic sign recognition systems in more realistic settings.

\section{Dataset and Classifier Description}

\subsection{LISA Traffic Sign Dataset}
The LISA Traffic Sign dataset is a widely used benchmark for evaluating traffic sign recognition models in real-world scenarios. It comprises over 7,800 annotated video frames captured from a vehicle-mounted camera in urban environments, encompassing 47 distinct classes of U.S. traffic signs, such as Stop, Speed Limit, Yield, and No U-Turn signs (Mogelmose et al., 2012). The dataset is particularly valued for its realistic driving conditions, including variations in illumination, weather, motion blur, camera angles, and partial occlusions, which more closely mimic the challenges faced by autonomous driving systems in real-time deployment.

Unlike synthetic or standardized datasets such as MNIST or GTSRB (German Traffic Sign Recognition Benchmark), the LISA dataset is characterized by significant intra-class variability and environmental noise, making it more challenging and better suited for evaluating robust deep learning models (Laroca et al., 2021). Images are typically resized to 128×128 pixels to meet the input dimension requirements of modern convolutional neural networks (CNNs), ensuring computational feasibility and compatibility with pretrained architectures.

For experimental consistency, the dataset is divided into training and testing subsets that preserve the original class distribution, which is crucial for preventing class imbalance biases and for conducting fair performance evaluations using metrics such as accuracy, precision, recall, and F1-score (Zhu et al., 2016). Several studies have used the LISA dataset to develop and benchmark approaches such as multi-scale CNNs, attention mechanisms, and domain adaptation techniques that aim to improve detection and classification accuracy in unconstrained environments.

\subsection{Convolutional Neural Network Classifier}
We employ a custom-designed convolutional neural network (CNN) architecture to perform multi-class classification over a dataset comprising 47 distinct traffic sign categories. CNNs have been widely adopted for image classification tasks due to their ability to automatically learn hierarchical feature representations from raw image data (LeCun et al., 1998; Krizhevsky et al., 2012). Our architecture is inspired by standard deep learning practices for traffic sign recognition (Ciresan et al., 2012; Sermanet, LeCun, 2011), and is deliberately kept compact to facilitate adversarial robustness analysis.

The network consists of three convolutional layers with increasing depth, employing 32, 64, and 128 filters respectively, each with a kernel size of 3×3. Every convolutional operation is followed by a Rectified Linear Unit (ReLU) activation function to introduce non-linearity, and a 2×2 max-pooling layer to downsample the feature maps, thereby reducing computational complexity and promoting spatial invariance. This progressively abstracts features from low-level edges and textures to higher-level semantic patterns, as observed in typical CNN hierarchies (Zeiler, Fergus, 2014).

The extracted features are then flattened and passed through two fully connected (dense) layers. The final layer outputs a 47-dimensional softmax probability vector, corresponding to the classification confidence scores for each traffic sign class. Training is carried out using the Adam optimizer (Kingma, Ba, 2015), with an initial learning rate of 0.001. The loss function used is categorical cross-entropy, a standard choice for multi-class classification problems.

To enhance convergence and model stability, input images are normalized across all RGB channels to have a mean of 0.5 and a standard deviation of 0.5. This normalization step ensures that input values are centered and scaled appropriately, which is known to accelerate training and reduce internal covariate shift (Ioffe, Szegedy, 2015).

This model serves as our baseline classifier and is subsequently used to assess the impact of adversarial perturbations in a controlled evaluation setting. We explore how imperceptible perturbations can degrade classification performance, a phenomenon first highlighted in the work of Szegedy et al. (2014) and further developed by Goodfellow et al. (2015) in the context of adversarial machine learning.

\section{Attack Methods}

\subsection{Fast Gradient Sign Method (FGSM)}
The Fast Gradient Sign Method (FGSM) is one of the most widely used adversarial attack techniques. Introduced by Goodfellow et al., FGSM generates adversarial examples by adding a perturbation to the original input image in the direction of the gradient of the loss with respect to the input. The adversarial example $x^{\text{adv}}$ is computed as:

\begin{equation}
x^{\text{adv}} = x + \epsilon \cdot \text{sign}(\nabla_x J(\theta, x, y))
\end{equation}

where $x$ is the original input, $y$ is the true label, $J(\theta, x, y)$ is the loss function, $\theta$ represents the model parameters and $\epsilon$ controls the magnitude of the perturbation. FGSM is a computationally efficient one-step method that aims to maximize loss by minimally moving the input in the adversarial direction.

\subsection{Projected Gradient Descent (PGD)}
Projected Gradient Descent (PGD) is a stronger iterative attack that generalizes FGSM by applying multiple small perturbation steps while keeping the adversarial examples within a specified $\epsilon$-ball around the original input. At each step, PGD updates the adversarial example as:

\begin{equation}
x_{t+1}^{\text{adv}} = \text{Proj}_{\epsilon}(x_t^{\text{adv}} + \alpha \cdot \text{sign}(\nabla_{x_t^{\text{adv}}} J(\theta, x_t^{\text{adv}}, y)))
\end{equation}

where $\alpha$ is the step size, and $\text{Proj}_{\epsilon}(\cdot)$ projects the updated adversarial example back onto the allowed $\epsilon$-ball centered at $x$. By iteratively applying small FGSM-like steps and projecting, PGD can generate more effective adversarial examples that are harder to defend against compared to single-step methods.

In our experiments, both FGSM and PGD attacks are performed under varying values of $\epsilon$ to study their effect on the classification accuracy of the traffic sign classifier.

\subsection{Visual Effect of Adversarial Attacks}

To illustrate the visual impact of adversarial perturbations, we present examples of how varying $\epsilon$ affects the appearance of traffic sign images under FGSM and PGD attacks.

\begin{figure}[htbp]
\centering
\includegraphics[width=0.48\textwidth]{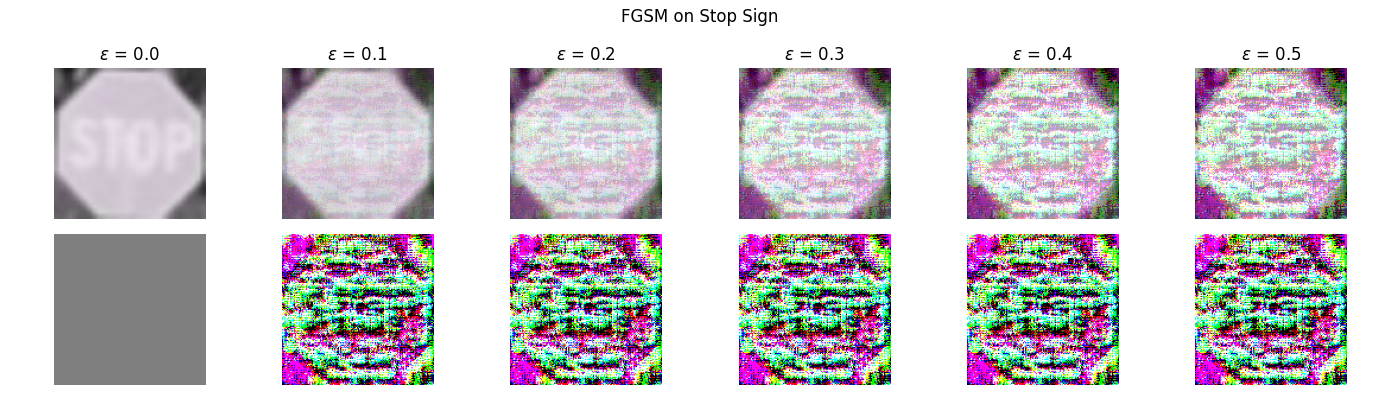}
\caption{FGSM attack on a Stop sign at increasing $\epsilon$ values. Top row: adversarial images. Bottom row: added perturbations.}
\label{fig:fgsm_grid}
\end{figure}

Figure~\ref{fig:fgsm_grid} shows adversarial examples of a Stop sign under the FGSM attack. As $\epsilon$ increases, the top row shows increasingly distorted adversarial images, while the bottom row displays the corresponding perturbations added to the original image.

\begin{figure}[htbp]
\centering
\includegraphics[width=0.48\textwidth]{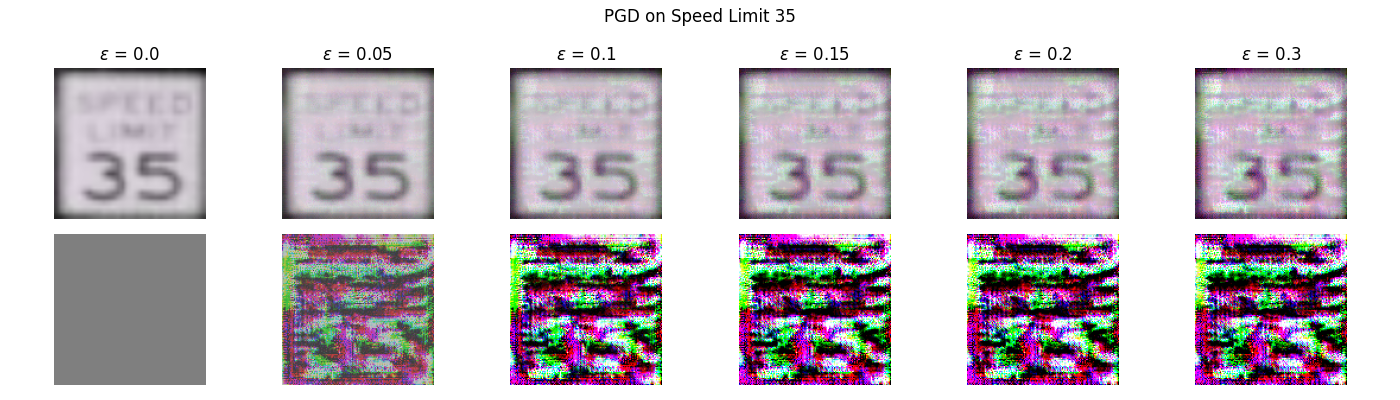}
\caption{PGD attack on a Speed Limit 35 sign at increasing $\epsilon$ values. Top row: adversarial images. Bottom row: added perturbations.}
\label{fig:pgd_grid}
\end{figure}

Similarly, Figure~\ref{fig:pgd_grid} illustrates the effect of PGD on a Speed Limit 35 sign. The attack introduces increasingly visible artifacts in the adversarial images as $\epsilon$ grows. The bottom row again highlights the perturbations, scaled for visual clarity.

\section{Experimental Setup and Results}

\subsection{Experimental Setup}
The convolutional neural network described in Section III was trained on the LISA Traffic Sign dataset using the Adam optimizer with a learning rate of 0.001 for 10 epochs. The input images were resized to $128\times128$ pixels and normalized to have zero mean and unit variance. The final model achieved a baseline classification accuracy of 84.93\% on the clean (unperturbed) test set.

To evaluate the model's robustness, adversarial examples were generated using both FGSM and PGD attacks at varying perturbation magnitudes ($\epsilon$). For FGSM, a single-step perturbation was applied, while for PGD, an iterative procedure with 10 steps and a step size of $\alpha=0.02$ was used. All attacks were performed in a white-box setting where the adversary has full access to the model's parameters.

\subsection{Results}
Tables \ref{tab:fgsm} and \ref{tab:pgd} summarize the model's classification accuracy under FGSM and PGD attacks, respectively, for different values of $\epsilon$.

\begin{table}[htbp]
\caption{Classification Accuracy Under FGSM Attack}
\renewcommand{\arraystretch}{1.3}
\centering
\normalsize
\begin{tabular}{|c|c|}
\hline
\textbf{Epsilon} & \textbf{Accuracy (\%)} \\
\hline
0.00 & 84.93 \\
0.10 & 17.26 \\
0.20 & 11.26 \\
0.30 & 9.48 \\
0.40 & 8.83 \\
0.50 & 7.86 \\
0.60 & 7.29 \\
\hline
\end{tabular}
\label{tab:fgsm}
\end{table}

\begin{table}[htbp]
\caption{Classification Accuracy Under PGD Attack}
\renewcommand{\arraystretch}{1.3}
\centering
\normalsize
\begin{tabular}{|c|c|}
\hline
\textbf{Epsilon} & \textbf{Accuracy (\%)} \\
\hline
0.00 & 84.93 \\
0.05 & 12.40 \\
0.10 & 5.19 \\
0.15 & 4.46 \\
0.20 & 4.21 \\
0.30 & 4.21 \\
\hline
\end{tabular}
\label{tab:pgd}
\end{table}

\subsection{Visualization of Attack Impact}
The degradation in classification accuracy as the perturbation magnitude increases is clearly observed in Tables \ref{tab:fgsm} and \ref{tab:pgd}. As the value of $\epsilon$ increases, the adversarial examples become more effective at fooling the classifier, leading to a sharp decline in accuracy. The FGSM attack shows a rapid drop in performance even at low $\epsilon$ values, while the PGD attack proves to be even more potent with consistent low accuracy across all tested perturbation levels.

\section{Conclusion}

In this paper, we replicated and extended adversarial attack methodologies, namely FGSM and PGD, on the LISA traffic sign classification dataset. Using a previously trained convolutional neural network (CNN) with 47 traffic sign classes, we demonstrated how single-step and iterative gradient-based perturbations significantly degrade classification accuracy as adversarial strength increases.

Our results show that even modest perturbations can reduce model accuracy from more than 84\% to less than 10\% for both FGSM and PGD, confirming the vulnerability of standard deep learning models to adversarial examples. Through both quantitative results and visual illustrations, we observed that increasing $\epsilon$ leads to distortions that become perceptible and eventually semantically misleading.

This study highlights the need for robust defense mechanisms in safety-critical applications such as autonomous driving. In future work, we will explore and evaluate various defense strategies, including adversarial training and input preprocessing techniques, to improve model resilience against such attacks.


\begin{thebibliography}{00}
\bibitem{b1} Szegedy, C., Zaremba, W., Sutskever, I., Bruna, J., Erhan, D., Goodfellow, I., Fergus, R. (2014) "Intriguing properties of neural networks.", \textit{arXiv preprint arXiv:1312.6199.}
\bibitem{b2} Goodfellow, I. J., Shlens, J., Szegedy, C. (2015) "Explaining and harnessing adversarial examples. International Conference on Learning Representations", \textit{ICLR}
\bibitem{b3} Madry et al. (2018) "Towards deep learning models resistant to adversarial attacks.", \textit{International Conference on Learning Representations (ICLR).}
\bibitem{b4} Athalye, A., Carlini, N., Wagner, D. (2018).  "Obfuscated gradients give a false sense of security: Circumventing defenses to adversarial examples.", \textit{nternational Conference on Machine Learning (ICML).}
\bibitem{b5} Cohen, J., Rosenfeld, E., Kolter, J. Z. (2019). "Certified adversarial robustness via randomized smoothing.", \textit{International Conference on Machine Learning (ICML).}
\bibitem{b6} Mogelmose, A., Trivedi, M. M.,  Moeslund, T. B. (2012). "Vision-based traffic sign detection and analysis for intelligent driver assistance systems: Perspectives and survey.", \textit{IEEE Transactions on Intelligent Transportation Systems, 13(4), 1484–1497.}
\bibitem{b7} Laroca, R., Severo, E., Zanlorensi, L. A., et al. (2021). "A robust real-time automatic license plate recognition based on the YOLO detector.", \textit{Journal of Real-Time Image Processing, 18(2), 505–519.}
\bibitem{b8} Zhu, Z., Liang, D., Zhang, S., Huang, X., Li, B.,  Hu, S. (2016). "Traffic-sign detection and classification in the wild. ", \textit{In Proceedings of the IEEE Conference on Computer Vision and Pattern Recognition (pp. 2110–2118).}
\bibitem{b9} LeCun, Y., Bottou, L., Bengio, Y.,  Haffner, P. (1998). "Gradient-based learning applied to document recognition.", \textit{Proceedings of the IEEE.}
\bibitem{b10} Krizhevsky, A., Sutskever, I.,  Hinton, G. E. (2012)."ImageNet classification with deep convolutional neural networks.", \textit{NeurIPS.}
\bibitem{b11} Ciresan, D. C., Meier, U., Masci, J.,  Schmidhuber, J. (2012)."Multi-column deep neural network for traffic sign classification.", \textit{Neural Networks.}
\bibitem{b12 } Sermanet, P.,  LeCun, Y. (2011)."Traffic sign recognition with multi-scale convolutional networks.", \textit{IJCNN.}

\bibitem{b13} Zeiler, M. D.,  Fergus, R. (2014). "Visualizing and understanding convolutional networks.", \textit{ECCV.}

\bibitem{b14} Kingma, D. P.,  Ba, J. (2015)."Adam: A method for stochastic optimization.", \textit{ICLR}

\bibitem{b15} Ioffe, S.,  Szegedy, C. (2015)."Batch normalization: Accelerating deep network training by reducing internal covariate shift.", \textit{ICML.}

\end{thebibliography}
\end{document}